\PassOptionsToPackage{unicode}{hyperref}
\PassOptionsToPackage{hyphens}{url}
\documentclass[
]{article}
\usepackage{xcolor}
\usepackage{amsmath,amssymb}
\usepackage{iftex}
\ifPDFTeX
  \usepackage[T1]{fontenc}
  \usepackage[utf8]{inputenc}
  \usepackage{textcomp} % provide euro and other symbols
\else % if luatex or xetex
  \usepackage{unicode-math} % this also loads fontspec
  \defaultfontfeatures{Scale=MatchLowercase}
  \defaultfontfeatures[\rmfamily]{Ligatures=TeX,Scale=1}
\fi
\usepackage{lmodern}
\ifPDFTeX\else
\fi
\IfFileExists{upquote.sty}{\usepackage{upquote}}{}
\IfFileExists{microtype.sty}{% use microtype if available
  \usepackage[]{microtype}
  \UseMicrotypeSet[protrusion]{basicmath} % disable protrusion for tt fonts
}{}
\makeatletter
\@ifundefined{KOMAClassName}{% if non-KOMA class
  \IfFileExists{parskip.sty}{%
    \usepackage{parskip}
  }{% else
    \setlength{\parindent}{0pt}
    \setlength{\parskip}{6pt plus 2pt minus 1pt}}
}{% if KOMA class
  \KOMAoptions{parskip=half}}
\makeatother
\usepackage{longtable,booktabs,array}
\usepackage{calc} % for calculating minipage widths
\usepackage{etoolbox}
\makeatletter
\patchcmd\longtable{\par}{\if@noskipsec\mbox{}\fi\par}{}{}
\makeatother
\IfFileExists{footnotehyper.sty}{\usepackage{footnotehyper}}{\usepackage{footnote}}
\makesavenoteenv{longtable}
\usepackage{graphicx}
\makeatletter
\newsavebox\pandoc@box
\newcommand*\pandocbounded[1]{% scales image to fit in text height/width
  \sbox\pandoc@box{#1}%
  \Gscale@div\@tempa{\textheight}{\dimexpr\ht\pandoc@box+\dp\pandoc@box\relax}%
  \Gscale@div\@tempb{\linewidth}{\wd\pandoc@box}%
  \ifdim\@tempb\p@<\@tempa\p@\let\@tempa\@tempb\fi% select the smaller of both
  \ifdim\@tempa\p@<\p@\scalebox{\@tempa}{\usebox\pandoc@box}%
  \else\usebox{\pandoc@box}%
  \fi%
}
\def\fps@figure{htbp}
\makeatother
\ifLuaTeX
  \usepackage{luacolor}
  \usepackage[soul]{lua-ul}
\else
  \usepackage{soul}
\fi
\usepackage{bookmark}
\IfFileExists{xurl.sty}{\usepackage{xurl}}{} % add URL line breaks if available
\hypersetup{
  pdftitle={An Analysis of Architectural Impact on LLM-based Abstract Visual Reasoning: A Systematic Benchmark on RAVEN-FAIR},
  hidelinks,
  pdfcreator={LaTeX via pandoc}}

\title{\textbf{An Analysis of Architectural Impact on LLM-based Abstract
Visual Reasoning: A Systematic Benchmark on RAVEN-FAIR}}
\author{}
\date{}

\begin{document}
\maketitle

\textbf{Authors:} Sinan Urgun, Seçkin Arı

Department of Computer Engineering, Faculty of Computer and Information
Sciences, Sakarya University, Sakarya, Turkey

Department of Computer Engineering, Faculty of Computer and Information
Sciences, Sakarya University, Sakarya, Turkey

\textbf{Corresponding Author:} sinan.urgun1@ogr.sakarya.edu.tr

\section{\texorpdfstring{\textbf{Abstract}}{Abstract}}\label{abstract}

This study aims to systematically evaluate the performance of large
language models (LLMs) in abstract visual reasoning problems. We
examined four LLM models (GPT-4.1-Mini, Claude-3.5-Haiku,
Gemini-1.5-Flash, Llama-3.3-70b) utilizing four different reasoning
architectures (single-shot, embedding-controlled repetition,
self-reflection, and multi-agent) on the RAVEN-FAIR dataset. Visual
responses generated through a three-stage process (JSON extraction, LLM
reasoning, and Tool Function) were evaluated using SSIM and LPIPS
metrics; Chain-of-Thought scores and error types (semantic
hallucination, numeric misperception) were analyzed. Results demonstrate
that GPT-4.1-Mini consistently achieved the highest overall accuracy
across all architectures, indicating a strong reasoning capability.
While the multi-agent architecture occasionally altered semantic and
numeric balance across models, these effects were not uniformly
beneficial. Instead, each model exhibited distinct sensitivity patterns
to architectural design, underscoring that reasoning effectiveness
remains model-specific. Variations in response coverage further emerged
as a confounding factor that complicates direct cross-architecture
comparison. To estimate the upper-bound performance of each
configuration, we report the best of five independent runs, representing
a best-case scenario rather than an averaged outcome. This multi-run
strategy aligns with recent recommendations {[}33{]}, which emphasize
that single-run evaluations are fragile and may lead to unreliable
conclusions.

\emph{\textbf{Keywords:} Abstract Visual Reasoning,
Raven\textquotesingle s Progressive Matrices, Large Language Models,
Chain-of-Thought, Self-Reflection, Multi-Agent Reasoning, RAVEN-FAIR,
Semantic Hallucination, Artificial Intelligence Evaluation}

\section{\texorpdfstring{\textbf{1.
Introduction}}{1. Introduction}}\label{introduction}

Abstract visual reasoning (AVR) is a key component of human
intelligence, often measured by tasks such as Raven\textquotesingle s
Progressive Matrices (RPM), which require discovering abstract rules
between visual concepts {[}1, 2{]}. For artificial intelligence,
mastering AVR is therefore considered a critical step toward achieving
human-like general intelligence {[}3{]}. Recent advances have been made
in RPM-style problems using large language models (LLMs) {[}4{]}.

Recent work regarding AVR has achieved high accuracy rates. However,
these methods (e.g., Valen solver) often included both the problem and
answer choices, requiring only the model to select the correct option
{[}5{]}. Furthermore, many deep learning solvers have incomprehensible
decision processes and unidentifiable error sources {[}6{]}. A
significant gap remains in benchmarks that systematically analyze
reasoning strategies {[}7, 8{]}, distinguish the source of errors (e.g.,
LLM vs. data format) {[}9, 10, 25{]}, or provide detailed error analysis
to understand inconsistencies {[}27{]}. This study addresses these gaps
by requiring LLMs to generate the answer *without* being provided any
answer choices. Four reasoning architectures, systematically benchmarked
and analyzed in detail, have been investigated.

In this study, we developed a system with a three-stage process (JSON
extraction, LLM reasoning, and Tool Function) pipeline to solve
RAVEN-FAIR problems. The system provides a comprehensive benchmark by
testing four reasoning architectures (single-shot, self-thought,
self-reflection, feature-based multi-agent) with four LLM models (Figure
1). Metrics such as accuracy, consistency, and error sources were
analyzed for each reasoning configuration, examining the effect of
reasoning strategy on success in detail. Thus, the objective was to
reveal the strengths and weaknesses of reasoning architectures in
LLM-based systems on challenging problems such as RAVEN-FAIR.

\includegraphics[width=5 in,height=2.16181in,alt={A diagram of a software }]{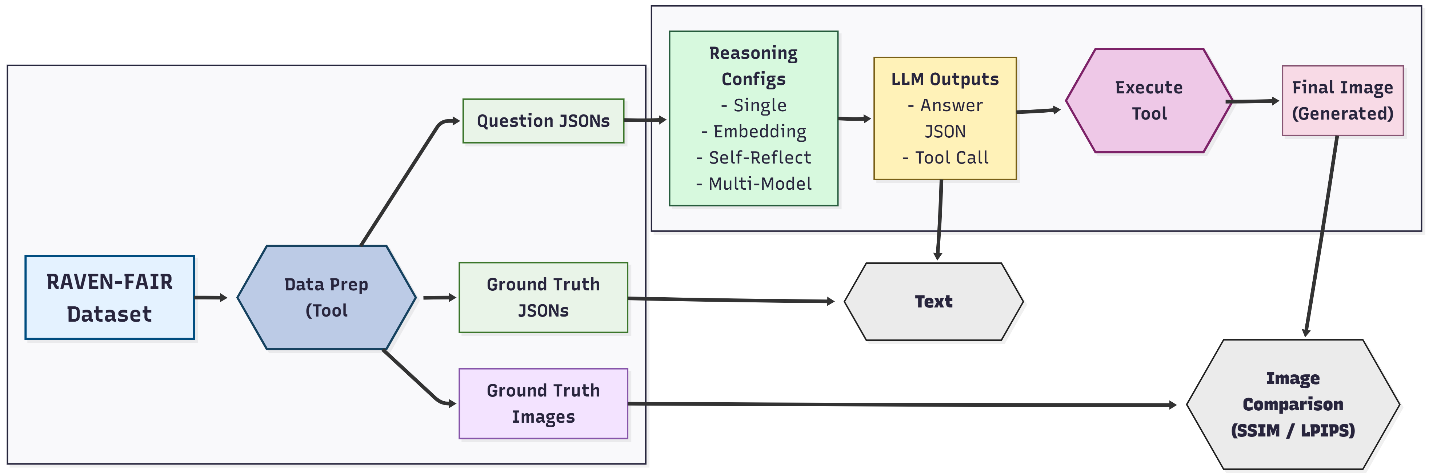}

\textbf{Figure 1.} This flow diagram illustrates the system architecture
of the study. The process begins with JSON extraction from data sources
and then branches into four different reasoning configurations. Each
model generates Chain-of-Thought (CoT) reasoning within its processing
steps and independently performs Tool Function calls. This function
produces the final visual/response output based on parameters determined
by the model. The generated results are analyzed in the evaluation stage
in terms of both visual similarity (SSIM/LPIPS) and textual consistency.

\section{\texorpdfstring{\textbf{2. Related Work}
}{2. Related Work }}\label{related-work}

In the original RAVEN dataset, abstract visual reasoning problems
contained "context-blind" errors that allowed models to correctly guess
answers based solely on the answer candidates, without considering the
context. This issue was addressed in the RAVEN-FAIR benchmark developed
by Benny et al., where distractor options were made context-sensitive,
thereby ensuring a fairer and more challenging reasoning test {[}12{]}.
This correction lays the essential foundation for fair and rigorous
evaluation.

In a comprehensive survey on deep learning--based solvers, Małkiński and
Mańdziuk {[}6{]} emphasized that many networks designed for solving RPM
and its derivatives operate as "black boxes," making it difficult to
determine whether errors originate from the data, the model, or the
architecture. The Funny-Valen-Tine study demonstrated that planning
solution distributions during the reasoning process can significantly
affect performance, and that the same model may produce contradictory
outcomes under different configurations {[}5{]}. The HriNet framework
showed that rules in RPM tasks can be learned hierarchically at the
panel, row, and matrix levels, thereby substantially improving accuracy
on balanced datasets {[}13{]}.

LLM-based approaches have also demonstrated that performance is strongly
influenced by inference strategies, such as multi-turn reasoning
{[}14{]}, output consistency optimization {[}15{]}, and
difficulty-graded training {[}16{]}.

Other effective AVR approaches include non-LLM methods, such as
multi-scale visual architectures for relational reasoning {[}12{]} and
even analysis of human eye-tracking data {[}17{]}.

\section{\texorpdfstring{\textbf{3.
Methodology}}{3. Methodology}}\label{methodology}

In our study, the LLM models were not provided with answer choices. The
models were required to analyze the question and generate the correct
answer solely based on their own constructed reasoning process. This
approach is crucial for evaluating whether the models truly possess
contextual reasoning capabilities. At the same time, it eliminates the
possibility of indirect inference by reasoning backward from the correct
answer, thereby ensuring a more transparent and reliable evaluation
process.

During the answer generation process, the models were required to
produce a visual output (in PNG format) through a predefined tool
function. The system\textquotesingle s evaluation relied not only on
textual outputs but also on results visually represented by this tool.

\subsection{\texorpdfstring{\textbf{3.1 Dataset
Preparation}}{3.1 Dataset Preparation}}\label{dataset-preparation}

A total of 1200 unique questions were generated using the official
GitHub repository of the RAVEN-FAIR benchmark {[}21{]}. This benchmark
provides problems in both \emph{.npz} and \emph{.xml} formats. For our
study, these data were converted into \emph{.png} images and JSON
formats in a 3×3 matrix arrangement through the custom Tool Function
developed in this work.

The Tool Function was designed to generate all combinations of shapes,
sizes, positions, angles, and styles supported by the RAVEN dataset,
according to externally provided parameters. Each problem consists of
nine sub-panels positioned as 1\_1, 1\_2, \ldots, 3\_3. Each sub-panel
was generated as an independent 160×160 image, which was then combined
to construct the complete 3×3 matrix. In the position of the missing
panel, a "?" symbol was inserted to indicate where the correct answer
must be determined.

\includegraphics[width=4.03163in,height=3.9587in,alt={.}]{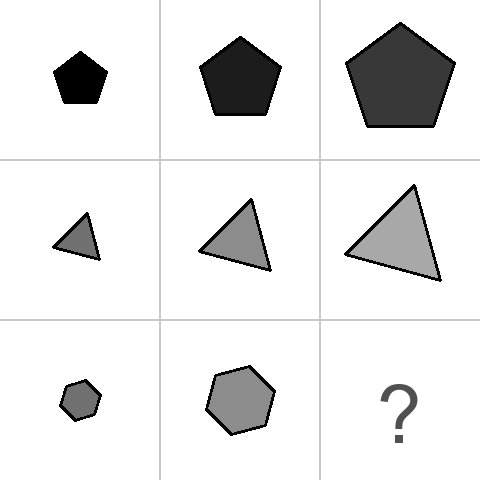}

\textbf{Figure 2.} Example dataset question. Each row in the figure
demonstrates variations in size and color of different geometric shapes.
The missing panel is located in the third row, third column.

The candidate answer panels were also generated at a resolution of 160 ×
160. This method ensures that each panel is drawn individually at a
fixed scale, thereby preventing potential semantic distortions caused by
drawing inconsistencies or scaling issues, and contributing to the
consistent generation of the entire question set.

In addition, JSON data representing each image was generated using the
generate\_visual\_panel() function. In this way, both visual and
structural data were obtained in a fully synchronized manner.

\subsection{\texorpdfstring{\textbf{3.2 Tool
Function}}{3.2 Tool Function}}\label{tool-function}

To transform LLM reasoning outputs into visual panels, a
generate\_visual\_panel() function was developed. This function,
detailed in our code repository, is a JSON-based API capable of
rendering all panel types, configurations, and attributes (e.g., shape,
color, size) supported by the RAVEN-FAIR benchmark at a 160x160
resolution.

\includegraphics[width=2 in,height=2 in,alt={.}]{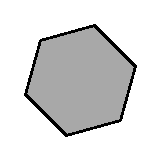}

\begin{verbatim}
"function_arguments": {
    "config_type": "singleton_center",
    "objects": [
        {
            "shape": "hexagon",
            "size": 0.8,
            "color": 3,
            "angle": 45
        }
    ]
}
\end{verbatim}

\textbf{Figure 3.} An example answer and tool function call generated by
GPT-4.1-Mini under the embedding-controlled configuration for Question
1.

\subsection{\texorpdfstring{\textbf{3.3 Reasoning
Configurations}}{3.3 Reasoning Configurations}}\label{reasoning-configurations}

In this study, four different reasoning architectures were evaluated on
a fixed 1200-question RAVEN-FAIR dataset {[}21{]}. These architectures
are single-shot reasoning, embedding-controlled repetition,
self-reflection, and feature-based multi-agent reasoning. Each
architecture introduces a distinct reasoning strategy into the
problem-solving process, aiming to reveal the decision-making quality,
error tolerance, and consistency of the LLMs.

\subsection{\texorpdfstring{\textbf{3.3.1 Single-Shot
Reasoning}}{3.3.1 Single-Shot Reasoning}}\label{single-shot-reasoning}

In this approach, the model is executed only once and directly generates
an answer based on the JSON representation of the problem. The answer
generation process does not include any feedback, correction,
repetition, or evaluation loops. This method is used to measure the
extent to which the LLM can solve the problem with its first intuitive
prediction.

Single-shot reasoning is a baseline for evaluating the performance
improvements achieved by more advanced strategies. Since the model is
required to capture the logical context in a single step, its tolerance
for error is low. However, as it does not include iterative feedback
loops, it is significantly faster in execution time and more efficient
in terms of system resources. This configuration provides a simple yet
effective testing ground for assessing how well the model can succeed on
its very first interaction with the prompt.

\subsection{\texorpdfstring{\textbf{3.3.2 Embedding-controlled
repetition}}{3.3.2 Embedding-controlled repetition}}\label{embedding-controlled-repetition}

In this architecture, the model generates an answer twice. The textual
embeddings of both answers are compared using cosine similarity. If the
similarity is below an experimentally determined threshold (0.8), the
response is considered unstable and regenerated to promote consistency.

\subsection{\texorpdfstring{\textbf{3.3.3
Self-Reflection}}{3.3.3 Self-Reflection}}\label{self-reflection}

This architecture assesses metacognitive abilities. After generating an
answer, the model initiates a reflective loop, assigning a confidence
score (on a scale of 0-10) to its own output. If this score is below a
threshold (8), the model deems the answer inadequate, revisits its
reasoning, and generates an alternative solution.

\subsection{\texorpdfstring{\textbf{3.3.4 Feature-Based Multi-Agent
Reasoning}}{3.3.4 Feature-Based Multi-Agent Reasoning}}\label{feature-based-multi-agent-reasoning}

This architecture provides a modular structure by assigning specialized
"agents" to analyze specific visual features (shape, color, position,
angle). Each agent provides a feature-specific output. A "master agent"
then integrates these outputs to construct the final JSON format that
calls the Tool Function, enabling more disentangled reasoning.

\subsection{\texorpdfstring{\textbf{3.4 Large Language
Models}}{3.4 Large Language Models}}\label{large-language-models}

In this study, four large language models (LLMs) with distinct
architectural designs and usage paradigms were selected to evaluate
their reasoning performance. The models tested are: GPT-4.1 Mini, Claude
3.5 Haiku, Gemini 1.5 Flash, and LLaMA 3.3 70B-q4.

The selection encompasses both commercial API-based systems and
open-source local models, thereby representing the diversity of modern
LLM ecosystems in terms of architecture and accessibility. GPT-4.1 Mini,
Claude 3.5, and Gemini 1.5 Flash were considered relatively more
accessible, cost-efficient solutions.

In contrast, LLaMA 3.3 70B-q4 was the only fully open-source and locally
deployed model included in this study. However, due to its high system
requirements, it cannot be evaluated as a cost-efficient alternative.
Instead, LLaMA was incorporated as a reference point to illustrate the
performance boundaries of open-source systems.

All models were tested using the same dataset and the same reasoning
architectures to perform an objective and comparative analysis of how
reasoning strategies vary depending on model architecture, access method
(API vs. local), and resource cost.

\textbf{Run Protocol.} To identify the upper-bound or
\textquotesingle peak\textquotesingle{} performance of each
architecture, rather than a statistical average, each configuration was
run five times. This approach mitigates the risk of reporting on a
single, anomalously poor run (n=1) while remaining feasible for a
systematic benchmark of this scale. Consistent with our goal of
evaluating peak capability, the best-performing run (by final-answer
accuracy) was preserved for analysis. This best-case reporting does not
capture run-to-run variance.

\section{\texorpdfstring{\textbf{4.
Evaluation}}{4. Evaluation}}\label{evaluation}

Even if a model generates the correct answer, the result cannot be
considered reliable if the Chain-of-Thought (CoT) process is shallow or
flawed. Conversely, if an incorrect answer is produced through a
detailed but logically invalid reasoning path, this indicates
higher-level errors such as semantic hallucination.

The evaluation assesses whether the model arrived at the correct answer,
how it arrived at it through its reasoning process, and why incorrect
answers were generated through flawed reasoning.

This approach is grounded in the increasingly prominent
\emph{"explainability-first evaluation"} methodologies in recent LLM
evaluation literature {[}20{]}.

This study employs a two-stage evaluation protocol:

\subsection{\texorpdfstring{\textbf{4.1 Final Answer
Accuracy}}{4.1 Final Answer Accuracy}}\label{final-answer-accuracy}

For each problem, every image produced by the LLM at the end of its
reasoning process is compared against the ground-truth answer. Two
widely used perceptual similarity metrics are employed:

\textbf{SSIM (Structural Similarity Index):} Compares luminance,
contrast, and structural information to yield a similarity score aligned
with human perception.

\textbf{LPIPS (Learned Perceptual Image Patch Similarity):} Uses deep
feature representations to assess perceptual differences between images,
providing a more human-aligned comparison {[}18{]}{[}19{]}.

For each LLM\textquotesingle s set of candidate outputs, the image with
the highest SSIM and lowest LPIPS is selected as the
model\textquotesingle s \emph{final answer}. If this answer matches the
ground truth, the prediction is counted as correct.

\subsection{\texorpdfstring{\textbf{4.2 Chain-of-Thought (CoT)
Evaluation}}{4.2 Chain-of-Thought (CoT) Evaluation}}\label{chain-of-thought-cot-evaluation}

Chain-of-Thought (CoT) denotes the step-by-step, justified reasoning
path by which an LLM arrives at its final answer. Prior work has shown
that such intermediate rationales can improve interpretability and
accuracy {[}20{]}. In this study, we evaluate not only the correctness
of the answer but also the quality of justification; rationale quality
is scored using a CoT Feature Score.

Scoring protocol. Each CoT output is scored on a 0--10 scale according
to the following criteria:

{\def\LTcaptype{none} % do not increment counter
\begin{longtable}[]{@{}
  >{\centering\arraybackslash}p{(\linewidth - 2\tabcolsep) * \real{0.1758}}
  >{\centering\arraybackslash}p{(\linewidth - 2\tabcolsep) * \real{0.8129}}@{}}
\toprule\noalign{}
\endhead
\bottomrule\noalign{}
\endlastfoot
\textbf{Score Range} & \textbf{Description} \\
\textbf{0} & No content; the model produced no CoT. \emph{(No
Content)} \\
\textbf{1--4} & Superficial, incomplete, or random explanations; no
coherent reasoning chain. \\
\textbf{5--7} & Moderate justification; rules are identified
heuristically, but complete consistency is lacking. \\
\textbf{8--10} & Detailed, explicit analysis of patterns and rules with
strong agreement with the result. \emph{(Perfect reasoning)} \\
\end{longtable}
}

\begin{itemize}
\item
  Scores ≥ 5 are labeled "Present" to indicate a substantive CoT
  process.
\item
  CoTs with a score of 0 are labeled "No Content."
\item
  Scores 1--4 indicate that content exists but is weak and unreliable.
\end{itemize}

\subsection{\texorpdfstring{\textbf{4.3 Answer Status: Definitions and
Importance}}{4.3 Answer Status: Definitions and Importance}}\label{answer-status-definitions-and-importance}

Each example is categorized into five distinct statuses based on the
correctness of the result and the quality of the CoT:

{\def\LTcaptype{none} % do not increment counter
\begin{longtable}[]{@{}
  >{\raggedright\arraybackslash}p{(\linewidth - 4\tabcolsep) * \real{0.1757}}
  >{\raggedright\arraybackslash}p{(\linewidth - 4\tabcolsep) * \real{0.3630}}
  >{\raggedright\arraybackslash}p{(\linewidth - 4\tabcolsep) * \real{0.4281}}@{}}
\toprule\noalign{}
\endhead
\bottomrule\noalign{}
\endlastfoot
\textbf{Status} & \textbf{Definition} & \textbf{Importance} \\
\textbf{perfect\_correct} & The correct answer is produced, and the CoT
is consistent, meaningful, and logical. & Demonstrates that the model
can generate both correct outputs and valid justifications. \\
\textbf{perfect\_wrong} & The CoT is structurally coherent, but the
final answer is incorrect. & May reveal errors such as semantic
hallucination or misleading pattern recognition. \\
\textbf{good\_correct} & The answer is correct, but the CoT is
incomplete, superficial, or partially heuristic. & Suggests that the
model sometimes guesses correctly or fails to explain its reasoning
adequately. \\
\textbf{poor\_wrong} & Both the CoT process and the final answer are
incorrect. & Indicates reasoning failures or numeric misperceptions. \\
\end{longtable}
}

\subsection{\texorpdfstring{\textbf{4.4 Error Types: Conceptual
Definitions and
Classification}}{4.4 Error Types: Conceptual Definitions and Classification}}\label{error-types-conceptual-definitions-and-classification}

In this study, incorrect cases are not only classified as "right/wrong"
but also analyzed according to specific error categories. These
categories are entirely derived from errors observed during experiments.

\subsubsection{\texorpdfstring{\textbf{4.4.1 Semantic
Hallucination}}{4.4.1 Semantic Hallucination}}\label{semantic-hallucination}

Semantic hallucination occurs when the model invents a nonexistent
pattern, rule, or feature and incorporates it into its Chain-of-Thought
(CoT) reasoning. This error is identified by an explicit mention of a
transformation in the CoT that does not exist in the ground truth---for
instance, inventing a "size decreasing" rule for a row where no size
change occurs. The primary \textbf{effect} of this error is that it
invalidates the entire reasoning process. Even if the model
coincidentally arrives at the correct answer, the justification is not
grounded in the provided data, which severely reduces the logical
reliability and trustworthiness of the output.

\subsubsection{\texorpdfstring{\textbf{4.4.2 Numeric
Misperception}}{4.4.2 Numeric Misperception}}\label{numeric-misperception}

~Numeric misperception refers to the misinterpretation or incorrect
transcription of numerical attributes such as size, angle, or color
value. This error is identified when a model\textquotesingle s response
differs from the ground truth only in its numerical values (e.g., a size
deviation ≥ 0.1). For instance, a model might write "0.3" in its
Chain-of-Thought for a shape whose actual size is 0.6. The primary
effect of this failure is that it prevents the model from constructing a
valid reasoning chain, as the foundational visual features are
misperceived from the outset.

\subsubsection{\texorpdfstring{\textbf{\hfill\break
4.4.3 Minor Deviation
(\emph{Perfect\_Wrong})}}{ 4.4.3 Minor Deviation (Perfect\_Wrong)}}\label{minor-deviation-perfect_wrong}

Cases where the overall reasoning is correct but minor deviations (e.g.,
color tone shifts, angle deviations ≤ ±45°, or size differences ≤ 0.1)
lead to an incorrect SSIM/LPIPS score. These are denoted as
Perfect\_Wrong.

\subsubsection{\texorpdfstring{\textbf{4.4.4 Missing or Invalid
Format}}{4.4.4 Missing or Invalid Format}}\label{missing-or-invalid-format}

Cases where the model\textquotesingle s JSON output fails, such as
missing mandatory fields (shape, color, etc.), using unsupported data
types, having a corrupted structure, or failing to call the Tool
Function.

\section{\texorpdfstring{\textbf{5. Results and
Discussion}}{5. Results and Discussion}}\label{results-and-discussion}

In this section, we analyze the performance of the four LLMs across the
four reasoning architectures. The results for each architecture are
presented sequentially, highlighting the key performance metrics,
model-specific analyses, and error patterns.

\subsection{\texorpdfstring{\textbf{5.1 Single-Shot Architecture
Results}}{5.1 Single-Shot Architecture Results}}\label{single-shot-architecture-results}

\subsubsection{\texorpdfstring{\hfill\break
\textbf{Table 1.} Performance comparison across models under the
single-shot
architecture\emph{.}}{ Table 1. Performance comparison across models under the single-shot architecture.}}\label{table-1.-performance-comparison-across-models-under-the-single-shot-architecture.}

{\def\LTcaptype{none} % do not increment counter
\begin{longtable}[]{@{}
  >{\raggedright\arraybackslash}p{(\linewidth - 8\tabcolsep) * \real{0.2321}}
  >{\raggedright\arraybackslash}p{(\linewidth - 8\tabcolsep) * \real{0.1798}}
  >{\raggedright\arraybackslash}p{(\linewidth - 8\tabcolsep) * \real{0.1844}}
  >{\raggedright\arraybackslash}p{(\linewidth - 8\tabcolsep) * \real{0.1874}}
  >{\raggedright\arraybackslash}p{(\linewidth - 8\tabcolsep) * \real{0.1844}}@{}}
\toprule\noalign{}
\endhead
\bottomrule\noalign{}
\endlastfoot
\emph{\textbf{MODEL}} & \emph{\textbf{GPT-4.1}}

\emph{\textbf{-Mini}} & \emph{\textbf{Claude-3.5}}

\emph{\textbf{-Haiku}} & \emph{\textbf{LLAMA-3.3}}

\emph{\textbf{-70b}} & \emph{\textbf{Gemini 1.5}}

\emph{\textbf{-Flash}} \\
\emph{\textbf{Coverage (\%)}} & \emph{\textbf{99.7\% (1196)}} &
\emph{\textbf{99.7\% (1196)}} & \emph{\textbf{98.0\% (1176)}} &
\emph{\textbf{97.0\% (1163)}} \\
\emph{\textbf{Success Rate (\%)}} & \emph{\textbf{46.9\% (561)}} &
\emph{\textbf{30.9\% (369)}} & \emph{\textbf{32.6\% (383)}} &
\emph{\textbf{34.9\% (406)}} \\
\emph{\textbf{Overall Accuracy (\%)(n=1200)}} & \emph{\textbf{46.91\%}}
& \emph{\textbf{30.8\%}} & \emph{\textbf{32.57\%}} &
\emph{\textbf{34.91\%}} \\
\emph{\textbf{Avg SSIM (Correct)}} & \emph{\textbf{0.946}} &
\emph{\textbf{0.918}} & \emph{\textbf{0.919}} & \emph{\textbf{0.925}} \\
\emph{\textbf{Avg LPIPS (Correct)}} & \emph{\textbf{0.086}} &
\emph{\textbf{0.124}} & \emph{\textbf{0.127}} & \emph{\textbf{0.119}} \\
\emph{\textbf{Perfect Wrong (\%)}} & \emph{\textbf{28.8\%}} &
\emph{\textbf{17.9\%}} & \emph{\textbf{18.2\%}} &
\emph{\textbf{23.3\%}} \\
\emph{\textbf{Poor Wrong (\%)}} & \emph{\textbf{71.2\%}} &
\emph{\textbf{82.1\%}} & \emph{\textbf{81.8\%}} &
\emph{\textbf{76.7\%}} \\
\emph{\textbf{CoT Score}} & \emph{\textbf{3.14}} & \emph{\textbf{8.08}}
& \emph{\textbf{8.31}} & \emph{\textbf{2.62}} \\
\end{longtable}
}

\textbf{Note.} All models were evaluated on 1,200 test questions.
Numbers in parentheses represent the number of successful trials.
Percentages for Perfect\_Wrong and Poor\_Wrong are calculated only over
incorrect answers. SSIM and LPIPS metrics are computed solely over
correctly answered questions.

Analysis of the single-shot results reveals distinct performance
profiles for each model. GPT-4.1-Mini led with the highest overall
accuracy (46.91\%) and the best perceptual similarity (SSIM 0.946),
demonstrating superior implementation capacity. Its strengths were high
coverage (99.7\%) and the lowest semantic hallucination (55.43\%), while
its primary weakness was a low CoT score (3.14), suggesting success
stemmed more from pattern matching than structured reasoning. In stark
contrast, LLaMA-3.3-70B and Claude-3.5-Haiku exemplified a "CoT-Accuracy
Paradox." LLaMA-3.3-70B achieved the highest CoT score (8.31) but only
moderate accuracy (32.57\%); one possible interpretation is that these
findings reflect a post-hoc rationalization effect in CoT evaluation
\textbf{{[}28{]}}. Similarly, Claude exhibited a high CoT score (8.08)
but low accuracy (30.85\%), characterized by a high semantic
hallucination rate (83.56\%). Finally, Gemini-1.5-Flash demonstrated
moderate accuracy (34.91\%) and ranked second best in terms of semantic
hallucination rate (76.76\%) and a low CoT score (2.62). These
variations align with findings that hallucination is deeply linked to
model architecture \textbf{{[}22{]}{[}23{]}{[}26{]}}.

\includegraphics[width=6 in,height=3.20903in,alt={.}]{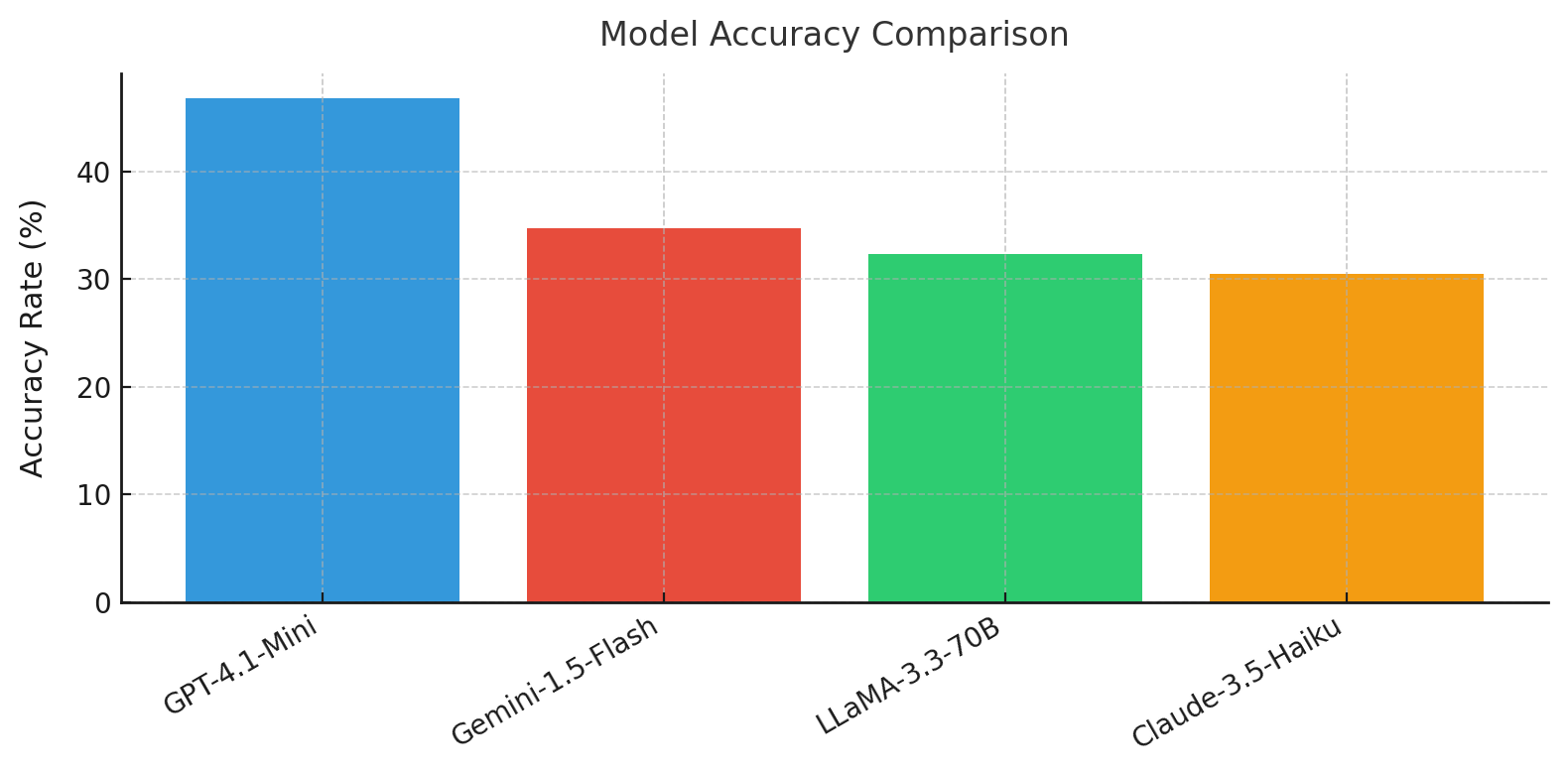}

\textbf{Figure 4. Performance comparison of models under the single-shot
architecture.} GPT-4.1-Mini leads with an accuracy of 46.91\%, while the
other models perform similarly within the 30--35\% range. This hierarchy
is generally preserved across all reasoning architectures.

\subsection{\texorpdfstring{\textbf{5.2 Embedding-Based Iterative
Architecture
Results}}{5.2 Embedding-Based Iterative Architecture Results}}\label{embedding-based-iterative-architecture-results}

\textbf{\hfill\break
Table 2.} \emph{Performance results across models under the
embedding-based iterative architecture.}

{\def\LTcaptype{none} % do not increment counter
\begin{longtable}[]{@{}
  >{\raggedright\arraybackslash}p{(\linewidth - 8\tabcolsep) * \real{0.2321}}
  >{\raggedright\arraybackslash}p{(\linewidth - 8\tabcolsep) * \real{0.1798}}
  >{\raggedright\arraybackslash}p{(\linewidth - 8\tabcolsep) * \real{0.1844}}
  >{\raggedright\arraybackslash}p{(\linewidth - 8\tabcolsep) * \real{0.1874}}
  >{\raggedright\arraybackslash}p{(\linewidth - 8\tabcolsep) * \real{0.1844}}@{}}
\toprule\noalign{}
\endhead
\bottomrule\noalign{}
\endlastfoot
\emph{\textbf{MODEL}} & \emph{\textbf{GPT-4.1}}

\emph{\textbf{-Mini}} & \emph{\textbf{Claude-3.5}}

\emph{\textbf{-Haiku}} & \emph{\textbf{LLAMA-3.3}}

\emph{\textbf{-70b}} & \emph{\textbf{Gemini 1.5}}

\emph{\textbf{-Flash}} \\
\emph{\textbf{Coverage (\%)}} & \emph{\textbf{99.8\% (1198)}} &
\emph{99.6\textbf{\% (}1195\textbf{)}} & \emph{\textbf{97.2\%
(}1166\textbf{)}} & \emph{\textbf{99.2\% (1190)}} \\
\emph{\textbf{Success Rate (\%)}} & \emph{53.9\textbf{\% (646)}} &
\emph{31.1\textbf{\% (372)}} & \emph{33.0\textbf{\% (}385\textbf{)}} &
\emph{33.7\textbf{\% (401)}} \\
\emph{\textbf{Overall Accuracy (\%)(n=1200)}} & \emph{\textbf{53.92\%}}
& \emph{\textbf{31.13\%}} & \emph{\textbf{33.02\%}} &
\emph{\textbf{33.70\%}} \\
\emph{\textbf{Avg SSIM (Correct)}} & \emph{\textbf{0.956}} &
\emph{\textbf{0.917}} & \emph{\textbf{0.924}} & \emph{\textbf{0.923}} \\
\emph{\textbf{Avg LPIPS (Correct)}} & \emph{\textbf{0.068}} &
\emph{\textbf{0.131}} & \emph{\textbf{0.118}} & \emph{\textbf{0.120}} \\
\emph{\textbf{Perfect Wrong (\%)}} & \emph{\textbf{33.5\%}} &
\emph{\textbf{20.3\%}} & \emph{\textbf{17.4\%}} &
\emph{\textbf{22.3\%}} \\
\emph{\textbf{Poor Wrong (\%)}} & \emph{\textbf{66.5\%}} &
\emph{\textbf{79.7\%}} & \emph{\textbf{82.6\%}} &
\emph{\textbf{77.7\%}} \\
\emph{\textbf{CoT Score}} & \emph{\textbf{3.73}} & \emph{\textbf{9.18}}
& \emph{\textbf{8.07}} & \emph{\textbf{1.90}} \\
\end{longtable}
}

The embedding-filtered architectures yielded mixed outcomes across
models. For GPT-4.1-Mini, accuracy increased markedly by +7.01 points to
53.92\%, mainly because the filter stabilized "Poor\_Wrong" responses
and reduced decision-layer drift. However, this mechanism could not
address deeper reasoning faults: semantic hallucination increased from
55.43\% to 71.01\%, indicating that semantic-level consistency does not
ensure numerical precision. The rise in ``Perfect\_Wrong'' cases from
169 to 182 further supports this observation.

In contrast, LLaMA-3.3-70B achieved only a +0.45\% gain, illustrating
the "consistency-in-error" phenomenon. The embedding filter repeatedly
endorsed identical mistakes because the model\textquotesingle s inherent
implementation weaknesses remained; this reveals a limitation of
self-consistency-based approaches, which may amplify flawed reasoning
{[}29{]}.

Claude-3.5-Haiku showed almost no improvement (+0.28\%) despite reaching
the highest chain-of-thought (CoT) score (9.18). This mismatch suggests
that self-consistency mechanisms {[}29{]} can produce an illusion of
stability in models lacking solid execution ability.

Gemini-1.5-Flash exhibited the opposite pattern: total accuracy dropped
(--1.21\%) and semantic hallucination up significantly (from 76.76\% to
83.56\%), numeric misperception increased (from 55.47\% to 70.25\%).
This change indicates that embedding-filtered architectures does not
guarantee improvements across all cases.

\subsection{\texorpdfstring{\textbf{5.3 Self-Reflection Architecture
Results}}{5.3 Self-Reflection Architecture Results}}\label{self-reflection-architecture-results}

\textbf{Table 3.} Performance results across models under the
self-reflection architecture.

{\def\LTcaptype{none} % do not increment counter
\begin{longtable}[]{@{}
  >{\raggedright\arraybackslash}p{(\linewidth - 8\tabcolsep) * \real{0.2321}}
  >{\raggedright\arraybackslash}p{(\linewidth - 8\tabcolsep) * \real{0.1798}}
  >{\raggedright\arraybackslash}p{(\linewidth - 8\tabcolsep) * \real{0.1844}}
  >{\raggedright\arraybackslash}p{(\linewidth - 8\tabcolsep) * \real{0.1874}}
  >{\raggedright\arraybackslash}p{(\linewidth - 8\tabcolsep) * \real{0.1844}}@{}}
\toprule\noalign{}
\endhead
\bottomrule\noalign{}
\endlastfoot
\emph{\textbf{MODEL}} & \emph{\textbf{GPT-4.1}}

\emph{\textbf{-Mini}} & \emph{\textbf{Claude-3.5}}

\emph{\textbf{-Haiku}} & \emph{\textbf{LLAMA-3.3}}

\emph{\textbf{-70b}} & \emph{\textbf{Gemini 1.5}}

\emph{\textbf{-Flash}} \\
\emph{\textbf{Coverage (\%)}} & \emph{\textbf{97.8\% (1173)}} &
\emph{75.2\textbf{\% (}902\textbf{)}} & \emph{\textbf{98.4\%
(}1181\textbf{)}} & \emph{\textbf{99.5\% (1194)}} \\
\emph{\textbf{Success Rate (\%)}} & \emph{47.4\textbf{\% (556)}} &
\emph{32.5\textbf{\% (293)}} & \emph{31.8\textbf{\% (}375\textbf{)}} &
\emph{33.2\textbf{\% (396)}} \\
\emph{\textbf{Overall Accuracy (\%)(n=1200)}} & \emph{\textbf{46.33\%}}
& \emph{\textbf{24.42\%}} & \emph{\textbf{31.25\%}} &
\emph{\textbf{33.0\%}} \\
\emph{\textbf{Avg SSIM (Correct)}} & \emph{\textbf{0.951}} &
\emph{\textbf{0.931}} & \emph{\textbf{0.925}} & \emph{\textbf{0.921}} \\
\emph{\textbf{Avg LPIPS (Correct)}} & \emph{\textbf{0.074}} &
\emph{\textbf{0.117}} & \emph{\textbf{0.118}} & \emph{\textbf{0.122}} \\
\emph{\textbf{Perfect Wrong (\%)}} & \emph{\textbf{33.5\%}} &
\emph{\textbf{21.8\%}} & \emph{\textbf{18.9\%}} &
\emph{\textbf{20.9\%}} \\
\emph{\textbf{Poor Wrong (\%)}} & \emph{\textbf{66.5\%}} &
\emph{\textbf{78.2\%}} & \emph{\textbf{81.1\%}} &
\emph{\textbf{79.1\%}} \\
\emph{\textbf{CoT Score}} & \emph{\textbf{2.99}} & \emph{\textbf{7.90}}
& \emph{\textbf{8.21}} & \emph{\textbf{3.36}} \\
\end{longtable}
}

Self-reflection mechanisms had divergent effects across models. For
GPT-4.1-Mini, accuracy slightly decreased (--0.58\%), while semantic
hallucination increased from 55.43\% to 81.20\%. This increase suggests
that reflection often rationalizes existing mistakes rather than
correcting them, thereby reinforcing confirmation bias rather than
encouraging exploratory reasoning.

LLaMA-3.3-70B also experienced a minor decline in performance
(--1.32\%). Although reflection reduced semantic errors by 1.19\%,
execution weaknesses persisted even with a high chain-of-thought (CoT)
score of 8.21, again emphasizing the gap between verbal rationalization
and actual problem-solving competence.

Claude-3.5-Haiku suffered the most severe degradation, with coverage
dropping drastically from 99.7\% to 75.2\% (a decrease of 24.5\%). This
loss fundamentally altered the evaluation sample, reflecting systematic
refusals and overly critical self-assessment, consistent with the
documented limitations of meta-cognitive strategies {[}10{]}.

Gemini-1.5-Flash showed a contrasting pattern: coverage increased from
97.0\% to 99.5\%, but accuracy declined by 1.91\%. Reflection failed to
stabilize semantic grounding, as hallucination increased from 76.76\% to
82.53\%. This effect was further reinforced by a rise in numeric
misperception (from 55.47\% to 61.90\%), highlighting the trade-off
between interpretive coherence and quantitative precision.

\subsection{\texorpdfstring{\textbf{5.4 Multi-Agent Architecture
Results}}{5.4 Multi-Agent Architecture Results}}\label{multi-agent-architecture-results}

\textbf{Table 4.} Performance results across models under the multi
agent architecture.

{\def\LTcaptype{none} % do not increment counter
\begin{longtable}[]{@{}
  >{\raggedright\arraybackslash}p{(\linewidth - 8\tabcolsep) * \real{0.2704}}
  >{\raggedright\arraybackslash}p{(\linewidth - 8\tabcolsep) * \real{0.1416}}
  >{\raggedright\arraybackslash}p{(\linewidth - 8\tabcolsep) * \real{0.1851}}
  >{\raggedright\arraybackslash}p{(\linewidth - 8\tabcolsep) * \real{0.1883}}
  >{\raggedright\arraybackslash}p{(\linewidth - 8\tabcolsep) * \real{0.1851}}@{}}
\toprule\noalign{}
\endhead
\bottomrule\noalign{}
\endlastfoot
\emph{\textbf{MODEL}} & \emph{\textbf{GPT-4.1}}

\emph{\textbf{Mini}} & \emph{\textbf{Claude-3.5}}

\emph{\textbf{Haiku}} & \emph{\textbf{LLAMA-3.3}}

\emph{\textbf{70b}} & \emph{\textbf{Gemini1.5}}

\emph{\textbf{Flash}} \\
\emph{\textbf{Coverage (\%)}} & \emph{\textbf{90.6\% (1087)}} &
\emph{\textbf{98.3\% (1180)}} & \emph{\textbf{97.8\% (1173)}} &
\emph{\textbf{98.9\% (1187)}} \\
\emph{\textbf{Success Rate (\%)}} & \emph{\textbf{49.8\% (541)}} &
\emph{\textbf{33.1\% (391)}} & \emph{\textbf{42.3\% (496)}} &
\emph{\textbf{30.6\% (363)}} \\
\emph{\textbf{Overall Accuracy (\%)(n=1200)}} & \emph{\textbf{45.08\%}}
& \emph{\textbf{32.58\%}} & \emph{\textbf{41.33\%}} &
\emph{\textbf{30.25\%}} \\
\emph{\textbf{Avg SSIM(Correct)}} & \emph{\textbf{0.929}} &
\emph{\textbf{0.929}} & \emph{\textbf{0.924}} & \emph{\textbf{0.909}} \\
\emph{\textbf{Avg LPIPS(Correct)}} & \emph{\textbf{0.119}} &
\emph{\textbf{0.123}} & \emph{\textbf{0.125}} & \emph{\textbf{0.164}} \\
\emph{\textbf{Perfect Wrong (\%)}} & \emph{\textbf{31.1\%}} &
\emph{\textbf{18.8\%}} & \emph{\textbf{18.6\%}} &
\emph{\textbf{15.5\%}} \\
\emph{\textbf{Poor Wrong (\%)}} & \emph{\textbf{68.9\%}} &
\emph{\textbf{81.2\%}} & \emph{\textbf{81.4\%}} &
\emph{\textbf{84.5\%}} \\
\emph{\textbf{CoT Score}} & \emph{\textbf{9.74}} & \emph{\textbf{9.01}}
& \emph{\textbf{9.40}} & \emph{\textbf{9.35}} \\
\end{longtable}
}

The multi-agent architecture produced distinct trade-offs across all
models. GPT-4.1-Mini attained the highest accuracy (45.08\%) and the
highest chain-of-thought (CoT) score (9.74), yet coverage decreased to
90.6\%, likely due to coordination overhead. This setup introduced a
significant drawback: numeric misperception increased slightly from
34.96\% to 45.05\%, indicating that greater reasoning depth did not
translate to improved quantitative reliability.

LLaMA-3.3-70B recorded its most significant improvement, with accuracy
increasing by +8.76 to 41.33\%. However, this gain came at a cost:
semantic hallucination increased from 83.23\% to 90.84\%, and numerical
reasoning degraded significantly, with misperception climbing from
66.08\% to 95.86\%. These results reinforce the observed
numeric--semantic trade-off.

Claude-3.5-Haiku experienced a notable decline in semantic consistency,
as hallucination increased slightly from 83.56\% to 85.68\%. In
contrast, its numerical reasoning improved, with misperception dropping
from 70.25\% to 55.39\%. This inverse relation between semantic and
quantitative performance further supports the conceptual framework of
modular reasoning strategies {[}13{]}.

Gemini-1.5-Flash exhibited a modest semantic improvement, as
hallucination decreased slightly from 76.76\% to 72.68\%. However, this
gain was offset by a notable decline in numerical reasoning
(misperception increasing from 55.47\% to 65.76\%), resulting in the
lowest overall accuracy (30.25\%) among all models under this
configuration.

\includegraphics[width=6 in,height=2.6in,alt={.}]{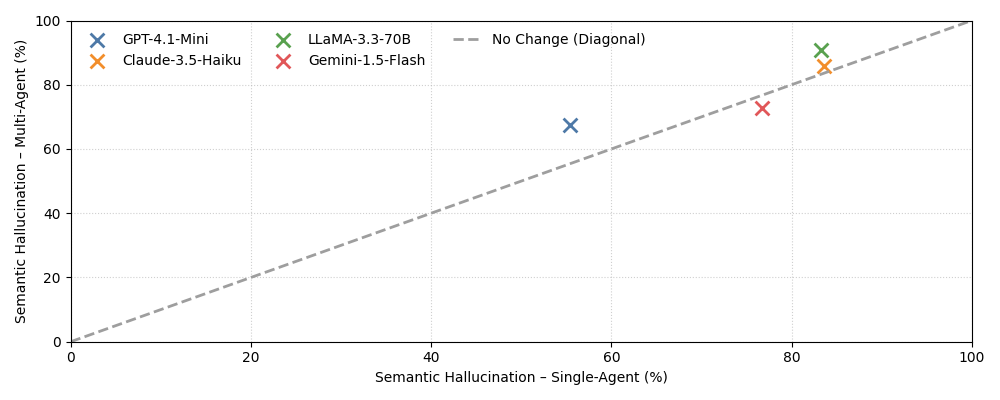}

\textbf{Figure 5.} Effect of the multi-agent architecture on semantic
hallucination. The diagonal line represents the "no change" reference.
Substantial improvements were observed in GPT-4.1-Mini (from 55.43\% to
67.40\%), while Claude-3.5-Haiku and Gemini-1.5-Flash showed only
minimal change. These results indicate that the multi-agent validation
did not consistently enhance semantic grounding.

\includegraphics[width=6 in,height=2.54583in,alt={.}]{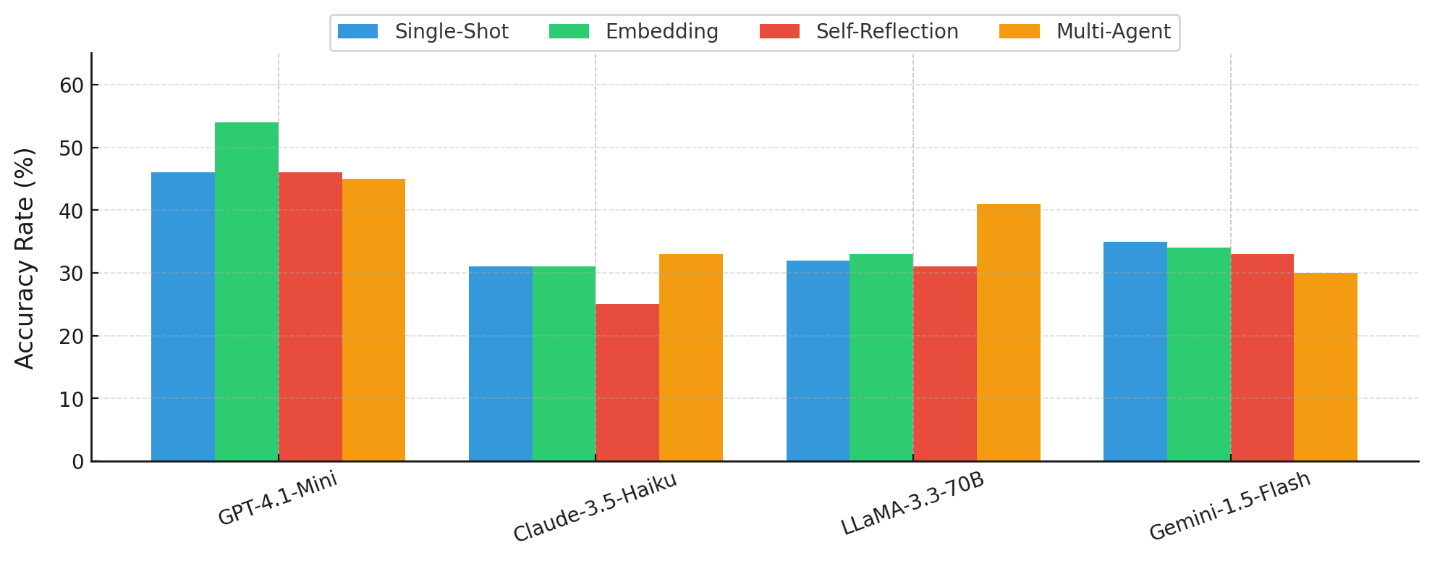}

\textbf{Figure 6.} Performance comparison of reasoning architectures.
The embedding-based architecture yielded a 7.01\% improvement for
GPT-4.1-Mini, while the multi-agent architecture achieved the most
significant performance gain with an 8.76\% increase for LLaMA-3.3-70B.
The self-reflection architecture had a minimal impact for most models
and led to a performance decline in Claude due to coverage loss.

\includegraphics[width=6 in,height=3.49514in,alt={.}]{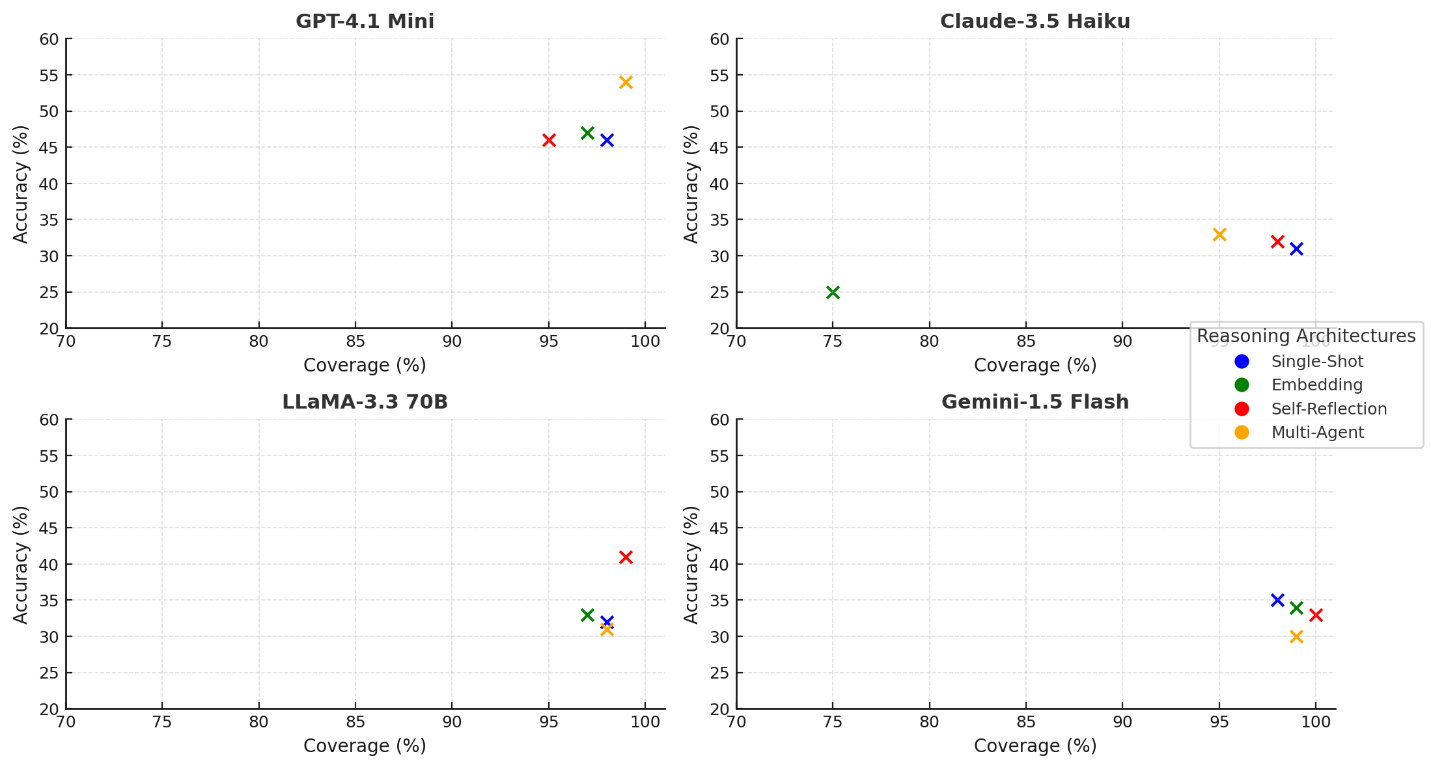}

\textbf{Figure 7.} Relationship between coverage rate and overall
accuracy. Bubble size represents the CoT score. The coverage drop to
75.2\% observed in Claude under the self-reflection architecture is
particularly noteworthy. This figure illustrates that coverage
variations act as a significant confounding factor, underscoring the
need for careful consideration of sample composition effects in
architecture comparison studies.

\includegraphics[width=6 in,height=3.86111in,alt={.}]{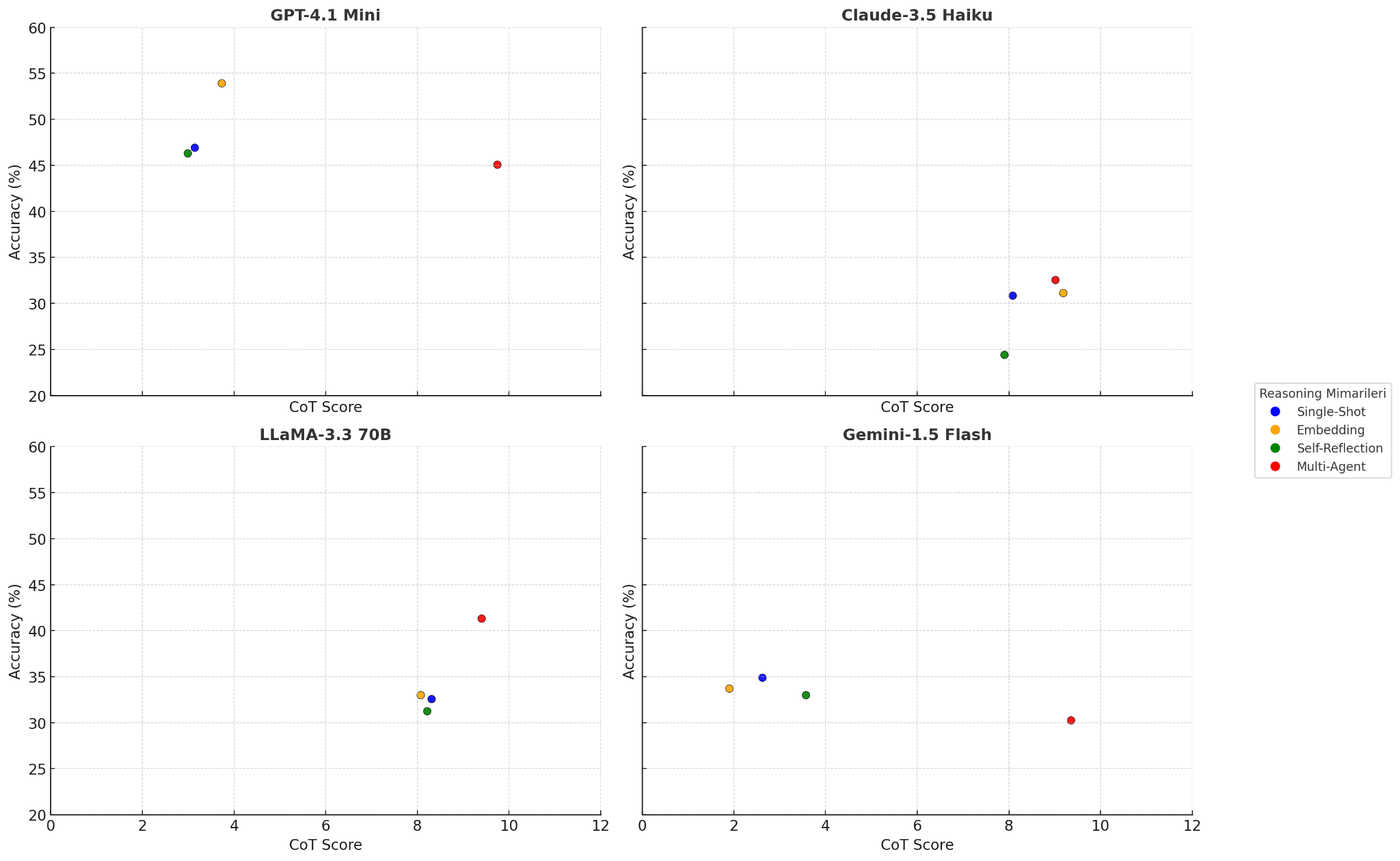}

\textbf{Figure 8.} Relationship between Chain-of-Thought (CoT) score and
actual performance. Bubble size represents coverage rate. The disconnect
between the high CoT scores of LLaMA-3.3-70B (83.1\%) and its moderate
accuracy (32.57\%) is clearly visible. This paradox highlights the
fundamental dissociation between explanation quality and execution
quality. One possible interpretation is that it reflects post-hoc
rationalization; however, further investigation would be required to
confirm this explanation.

\includegraphics[width=6 in,height=3.19861in,alt={.}]{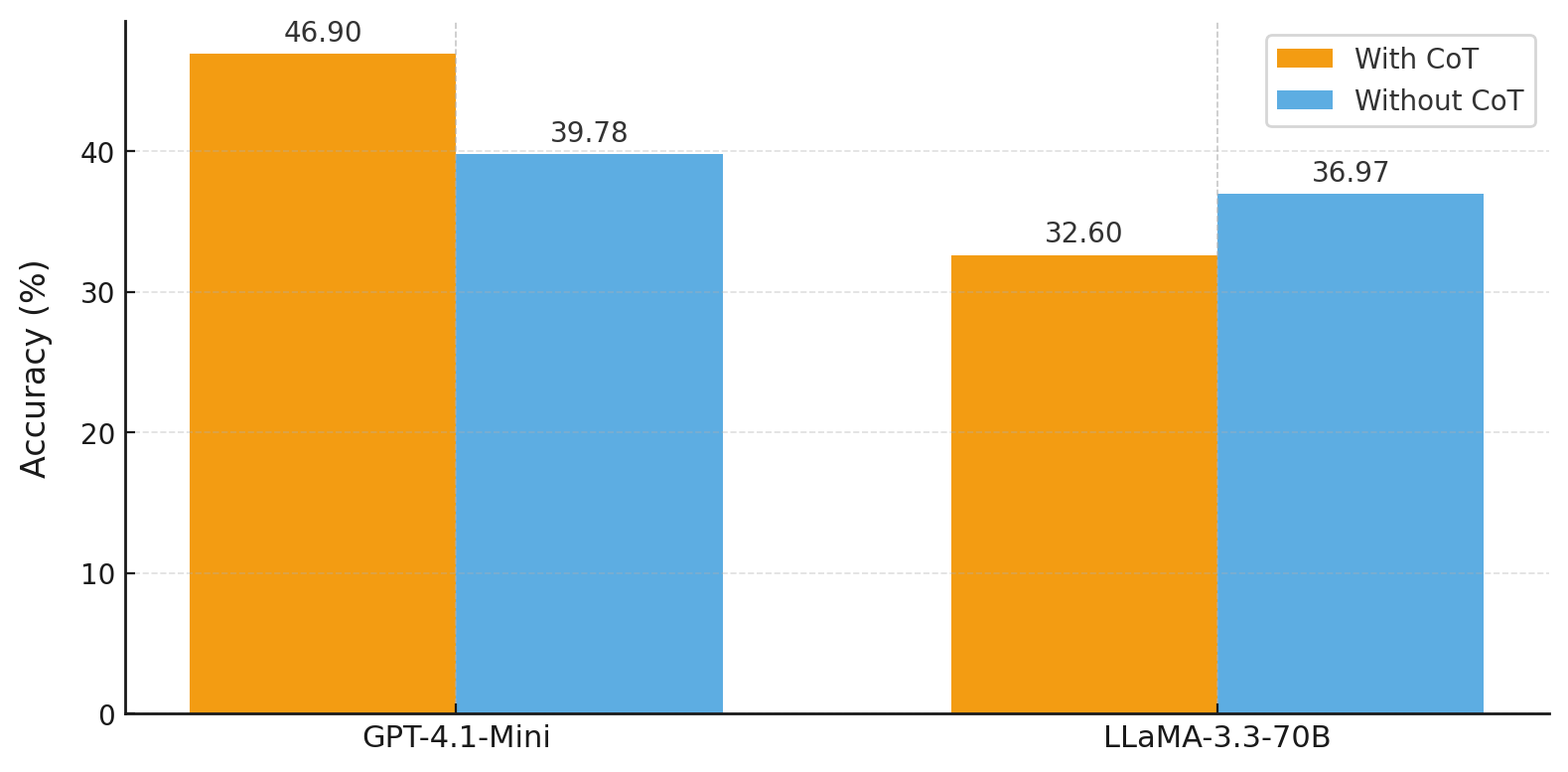}

\textbf{Figure 9.} Effect of CoT Presence on Model Performance (Post-Hoc
Analysis). The chart illustrates the performance decrease for
GPT-4.1-Mini and the performance increase for LLaMA-3.3-70B when
Chain-of-Thought (CoT) instructions are removed. This finding supports
the hypothesis that CoT effectiveness is model-specific.

\textbf{Note:} This divergence is hypothesized to stem from the
qualitative differences in CoT approaches.
GPT-4.1-Mini\textquotesingle s concise reasoning appears beneficial,
whereas LLaMA-3.3-70B\textquotesingle s overly detailed and complex CoTs
may introduce ambiguity and act as a hindrance.

\section{\texorpdfstring{\textbf{6. Conclusion and
Discussion}}{6. Conclusion and Discussion}}\label{conclusion-and-discussion}

\subsection{\texorpdfstring{\textbf{6.1 Key Findings and
Contributions}}{6.1 Key Findings and Contributions}}\label{key-findings-and-contributions}

This study systematically evaluated four reasoning architectures on the
RAVEN-FAIR benchmark, demonstrating that architectural selection is
critical and that performance patterns are model-specific. GPT-4.1-Mini
consistently achieved the highest accuracy across all configurations.
Our findings highlight three primary contributions:

\begin{enumerate}
\def\labelenumi{\arabic{enumi}.}
\item
  \textbf{The CoT--Performance Dissociation:} We found a systematic
  disconnect between explanation quality and execution quality (the
  "CoT-Accuracy Paradox"). High CoT scores (e.g., LLaMA-3.3-70B) did not
  predict high accuracy, and vice versa, supporting prior observations
  on the challenges of CoT evaluation \textbf{{[}23{]}}.
\item
  \textbf{The Semantic--Numeric Trade-off:}
\end{enumerate}

\begin{quote}
Contrary to earlier assumptions, the multi-agent architecture did not
uniformly improve semantic grounding. For instance, hallucination
slightly increased in Claude-3.5-Haiku (83.56 → 85.68\%) and
Gemini-1.5-Flash (76.76 → 72.68\%), while numeric misperception in some
models (e.g., Claude 70.25 → 55.39\%) showed moderate improvement. These
mixed effects suggest that multi-agent validation alters reasoning
balance rather than universally enhancing semantic precision.
\end{quote}

\begin{enumerate}
\def\labelenumi{\arabic{enumi}.}
\setcounter{enumi}{2}
\item
  \textbf{Coverage Bias as a Confounding Factor:} We identified coverage
  variation (ranging from 75.2\% to 99.8\%) as a major methodological
  confounder. This was most evident in the self-reflection architecture,
  where a 24.5\% coverage loss in Claude invalidated performance
  comparisons and suggested over critical behavior, aligning with
  limitations of self-consistency approaches \textbf{{[}29{]}}.
\end{enumerate}

\subsection{\texorpdfstring{\textbf{6.2 Practical
Implications}}{6.2 Practical Implications}}\label{practical-implications}

For applications with limited computational resources, GPT-4.1-Mini
combined with the single-shot architecture may provide an optimal
cost--performance balance. For tasks where semantic grounding is
critical, multi-agent architectures appear promising; however, when high
numeric precision is required, our findings show that additional
coordination mechanisms are necessary. The failures of self-reflection
suggest its parameters require careful tuning.

\subsection{\texorpdfstring{\textbf{6.3 Methodological Limitations and
Future
Work}}{6.3 Methodological Limitations and Future Work}}\label{methodological-limitations-and-future-work}

This study is constrained by its reliance on best-run reporting rather
than averaged statistics, which prevents variance analysis and
statistical significance testing. Future research must address this by
(1) enhancing statistical rigor with average performance, confidence
intervals, and significance testing; (2) developing hybrid architectures
to mitigate the identified trade-offs, such as combining multi-agent
validation with embedding consistency; and (3) addressing the root
causes of coverage loss, such as tool-function calling failures and poor
multi-agent coordination (e.g., "size agent" errors).

\subsection{\texorpdfstring{\textbf{6.4 Data and Code
Availability}}{6.4 Data and Code Availability}}\label{data-and-code-availability}

The source code and dataset generated or analyzed during this study are
available at:

GitHub Repository (LLM codes):
\href{https://github.com/SinanUrgunWork/An-Analysis-of-Architectural-Impact-on-LLM-based-Abstract-Visual-Reasoning/tree/main}{\ul{{[}link{]}}}{[}31{]}

Google Drive (Data):
\href{https://drive.google.com/drive/folders/1Q_YRu5hCFw2cVMATqI_WBwiTSgPSKbk9?usp=sharing}{\ul{{[}link{]}}}{[}32{]}

\subsection{\texorpdfstring{\textbf{6.5
Funding}}{6.5 Funding}}\label{funding}

This research did not receive any specific grant from funding agencies
in the public, commercial, or not-for-profit sectors.

\subsection{\texorpdfstring{\textbf{6.6 Declaration of Competing
Interests}}{6.6 Declaration of Competing Interests}}\label{declaration-of-competing-interests}

The authors declare no competing financial or personal interests.

\subsection{\texorpdfstring{\textbf{6.7
Acknowledgements}}{6.7 Acknowledgements}}\label{acknowledgements}

The authors acknowledge the use of AI-assisted tools in the Preparation
of this manuscript. Specifically, OpenAI\textquotesingle s GPT and
Anthropic\textquotesingle s Claude were employed to generate
illustrative figures, support translation and grammar correction, and
draft prompts used in coding tasks. All outputs were carefully reviewed
and validated by the authors, who take full responsibility for the
scientific content and conclusions presented in this paper.

\section{\texorpdfstring{\textbf{References}
}{References }}\label{references}

{[}1{]} J.C. Raven, J.H. Court, Raven\textquotesingle s Progressive
Matrices, H.K. Lewis \& Co Ltd, London, 1938.

{[}2{]} D. Barrett, F. Hill, A. Santoro, A. Morcos, T. Lillicrap,
Measuring abstract reasoning in neural networks, in: Proceedings of the
35th International Conference on Machine Learning, ICML 2018, pp.
511-520.

{[}3{]} W. Zhang, F. Gao, B. Shi, H. Lu, RAVEN: A dataset for relational
and analogical visual reasoning, in: Proceedings of the IEEE Conference
on Computer Vision and Pattern Recognition, CVPR 2019, pp. 5317-5327.

{[}4{]} F. Shi, B. Li, X. Xue, Beyond task-specific reasoning: A unified
conditional generative framework for abstract visual reasoning, in:
Proceedings of the 38th International Conference on Machine Learning,
ICML 2025, pp. 8847-8857.

\textbf{{[}5{]}} R. Song, X. Liu, Q. Zheng, et al., "Funny Valen-Tine:
Solving visual abstract reasoning problems through defining the solution
distribution," arXiv preprint arXiv:2407.02688, 2024.

\textbf{{[}6{]}} M. Małkiński and J. Mańdziuk, "Deep learning methods
for abstract visual reasoning: A survey on Raven\textquotesingle s
Progressive Matrices," ACM Computing Surveys, vol. 57, no. 7, pp. 1-36,
2025.

{[}7{]} J. Smith, K. Lee, T. Wang, Inference-time computations for LLM
reasoning: Sys2Bench benchmark, arXiv preprint arXiv:2402.11234, 2024.

{[}8{]} X. Li, Y. Zhang, S. Chen, et al., R-Bench: Graduate-level
multi-disciplinary benchmarks for LLM \& MLLM complex reasoning
evaluation, arXiv preprint arXiv:2403.15678, 2025.

\textbf{{[}9{]}} Q. Wang, Z. Wang, Y. Su, H. Tong, and Y. Song,
"Rethinking the bounds of LLM reasoning: Are multi-agent discussions the
key?" arXiv preprint arXiv:2402.18272, 2024.

\textbf{{[}10{]}} A. Havrilla, S. Raparthy, C. Nalmpantis, et al.,
"GLoRe: When, where, and how to improve LLM reasoning via global and
local refinements," arXiv preprint arXiv:2402.10963, 2024.

{[}11{]} L. Zhou, H. Sun, J. Wu, et al., KG-LLM-Bench: A scalable
benchmark for evaluating LLM reasoning on textualized knowledge graphs,
arXiv preprint arXiv:2406.17890, 2025.

\textbf{{[}12{]}} Y. Benny, N. Pekar, and L. Wolf, "Scale-localized
abstract reasoning," in Proceedings of the IEEE/CVF Conference on
Computer Vision and Pattern Recognition (CVPR), 2021, pp. 12557-12565.

\textbf{{[}13{]}} F. Shi, B. Li, and X. Xue, "Hierarchical Rule
Induction Network for Abstract Visual Reasoning," arXiv preprint
arXiv:2002.06838, 2020.

\textbf{{[}14{]}} X. Tian, et al., "Think Twice: Enhancing LLM reasoning
by scaling multi-round test-time thinking," arXiv preprint
arXiv:2503.19855, 2025.

\textbf{{[}15{]}} Q. Zhang, H. Wu, C. Zhang, P. Zhao, and Y. Bian,
"Right question is already half the answer: Fully unsupervised LLM
reasoning incentivization," arXiv preprint arXiv:2504.05812, 2025.

{[}16{]} DeepDistill, Difficulty-graded data training for LLM reasoning,
arXiv preprint arXiv:2409.20123, 2025.

{[}17{]} S. Ma, N. Jia, Measuring Raven\textquotesingle s Progressive
Matrices combining eye-tracking technology and machine learning models,
Journal of Intelligence 12(1) (2024) 116.

{[}18{]} Z. Wang, A.C. Bovik, H.R. Sheikh, E.P. Simoncelli, Image
quality assessment: from error visibility to structural similarity, IEEE
Transactions on Image Processing 13(4) (2004) 600-612.

{[}19{]} R. Zhang, P. Isola, A.A. Efros, E. Shechtman, O. Wang, The
Unreasonable Effectiveness of Deep Features as a Perceptual Metric, in:
Proceedings of the IEEE Conference on Computer Vision and Pattern
Recognition, CVPR 2018, pp. 586-595.

{[}20{]} B. Cheng, Q. Wu, T. Liu, A. van den Hengel, Measuring
Robustness of Visual Question Answering Models, arXiv preprint
arXiv:2104.00587, 2021.

{[}21{]} Y. Benny, RAVEN\_FAIR: A fair benchmark for RPM tasks, GitHub
repository, https://github.com/yanivbenny/RAVEN\_FAIR, Accessed: 7
Ağustos 2025.

{[}22{]} Li, Y., Du, Y., Zhou, K., Wang, J., Zhao, W. X., \& Wen, J.-R.
(2023). Evaluating Object Hallucination in Large Vision-Language Models
(POPE)\textbf{.} \emph{EMNLP 2023.}

{[}23{]} Zhang, Z., Zhang, A., Li, M., Zhao, H., Karypis, G., \& Smola,
A. (2024). Multimodal Chain-of-Thought Reasoning in Language
Models\textbf{.} \emph{TMLR, May 2024.}

{[}24{]} Guo, C., Pleiss, G., Sun, Y., \& Weinberger, K. Q. (2017). On
Calibration of Modern Neural Networks\textbf{.} \emph{ICML 2017 (PMLR
70).}

{[}25{]} Sweller, J. (2011). Cognitive Load Theory\textbf{.}
\emph{Psychology of Learning and Motivation}, 55, 37--76. Academic
Press.

{[}26{]} Bai, Z., Wang, P., Xiao, T., et al. (2024). Hallucination of
Multimodal Large Language Models: A Survey. \emph{arXiv:2404.18930.}

{[}27{]} Lakshminarayanan, B., Pritzel, A., \& Blundell, C. (2017).
Simple and Scalable Predictive Uncertainty Estimation using Deep
Ensembles. In \emph{NeurIPS 2017}.

{[}28{]} Turpin, M., Michael, J., Perez, E., \& Bowman, S. R. (2023).
Language Models Don\textquotesingle t Always Say What They Think.
arXiv:2305.04388.

{[}29{]} Wang, X., Wei, J., Schuurmans, D., et al. (2022).
Self-Consistency Improves Chain of Thought Reasoning in Language Models.
arXiv:2203.11171.

{[}30{]} Y. Geifman, R. El-Yaniv, Selective classification for deep
neural networks, in: Advances in Neural Information Processing Systems
(NeurIPS), 2017.

{[}31{]} S. Urgun, An Analysis of Architectural Impact on LLM-based
Abstract Visual Reasoning (GitHub Repository: LLM codes), Available at:
https://github.com/SinanUrgunWork/An-Analysis-of-Architectural-Impact-on-LLM-based-Abstract-Visual-Reasoning/tree/main
(accessed: 09/08/2025).

{[}32{]} S. Urgun, An Analysis of Architectural Impact on LLM-based
Abstract Visual Reasoning (Google Drive: Data), Available at:
https://drive.google.com/drive/folders/1Q\_YRu5hCFw2cVMATqI\_WBwiTSgPSKbk9?usp=sharing,
accessed: 09/08/2025.

{[}33{]} M.A. Alvarado Gonzalez, M. Bruno Hernandez, M.A. Peñaloza
Perez, B. Lopez Orozco, J.T. Cruz Soto, S. Malagon, DO REPETITIONS
MATTER? STRENGTHENING RELIABILITY IN LLM EVALUATIONS, arXiv preprint
arXiv:2509.24086, 2025.

\end{document}